\documentclass[runningheads]{llncs}
\usepackage{graphicx}
\usepackage{todonotes}
\usepackage{amssymb}
\usepackage[colorlinks = true,
            linkcolor = black,
            urlcolor  = blue,
            citecolor = black,
            anchorcolor = black]{hyperref}
\usepackage{booktabs}
\usepackage{multirow}
\usepackage{gensymb}
\usepackage{amsmath}
\usepackage{lscape}
\usepackage{subcaption}

\usepackage{academicons}
\usepackage{xcolor}

\newcommand{\orcid}[1]{\href{https://orcid.org/#1}{\textcolor[HTML]{A6CE39}{\aiOrcid}}}

\newcommand{\customfootnotetext}[2]{{
  \renewcommand{\thefootnote}{#1}
  \footnotetext[0]{#2}}}
\begin{document}
\title{Using Soft Labels to Model Uncertainty in Medical Image Segmentation}
\titlerunning{Using Soft Labels to Model Uncertainty in Medical Image Segmentation}
%
\author{João Lourenço Silva \href{https://orcid.org/0000-0001-6475-3042}{\includegraphics[scale=0.6]{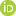}} \\
Arlindo L. Oliveira \href{https://orcid.org/0000-0001-8638-5594}{\includegraphics[scale=0.6]{Images/ORCIDiD_icon16x16.png}}}

\authorrunning{J. L. Silva and A. L. Oliveira}
%
\institute{Instituto Superior Técnico / INESC-ID
\email{\{joao.lourenco.silva,arlindo.oliveira\}@tecnico.ulisboa.pt}}
\maketitle              
\begin{abstract}
Medical image segmentation is inherently uncertain. For a given image, there may be multiple plausible segmentation hypotheses, and physicians will often disagree on lesion and organ boundaries. To be suited to real-world application, automatic segmentation systems must be able to capture this uncertainty and variability. Thus far, this has been addressed by building deep learning models that, through dropout, multiple heads, or variational inference, can produce a set - infinite, in some cases - of plausible segmentation hypotheses for any given image. However, in clinical practice, it may not be practical to browse all hypotheses. Furthermore, recent work shows that segmentation variability plateaus after a certain number of independent annotations, suggesting that a large enough group of physicians may be able to represent the whole space of possible segmentations. Inspired by this, we propose a simple method to obtain soft labels from the annotations of multiple physicians and train models that, for each image, produce a single well-calibrated output that can be thresholded at multiple confidence levels, according to each application's precision-recall requirements. We evaluated our method on the MICCAI 2021 QUBIQ challenge, showing that it performs well across multiple medical image segmentation tasks, produces well-calibrated predictions, and, on average, performs better at matching physicians' predictions than other physicians.

\end{abstract}

\section{Introduction}

Accurate segmentation of medical images is crucial in diagnosing and planning the treatment of multiple pathologies. Nevertheless, it is also very laborious and time-consuming, spurring great interest in the development of automatic segmentation mechanisms.

In the last few years, deep learning systems have achieved high performance in the segmentation of several organs and anatomical structures \cite{lei2020medical}. However, most methods do not account for the uncertainty inherent to these tasks. For a given image, there may be multiple plausible segmentations, and physicians will often disagree on the zones of interest and their contours. Thus, models should be able to capture uncertainty and express it in their predictions. Otherwise, they risk biasing physicians, which may lead to misdiagnosis and sub-optimal treatment.

To date, most work on uncertainty estimation in medical image segmentation focuses on being able to produce multiple plausible outputs for a given image \cite{baumgartner2019phiseg,hu2019supervised,kohl2019hierarchical,kohl2018probabilistic,monteiro2020stochastic}. However, in clinical practice, it may impractical to browse all hypotheses. Furthermore, recent research \cite{joskowicz2019inter} shows that, even though segmentation variability increases with the number of annotators, it plateaus after a certain data and task-dependent number of independent annotations, implying that, although multiple plausible segmentations exist for a given input, they can be encompassed by the annotations of a sufficiently large group of physicians.

In this work, we follow a trend orthogonal to that of previous work. Rather than aiming to build probabilistic models that can produce various plausible hypotheses, we propose to train deterministic models on soft labels built from the annotations of multiple physicians. We evaluated our method on datasets from the MICCAI 2021 QUBIQ challenge. The results showed that it performs well compared to alternative approaches and produces well-calibrated outputs across a range of medical image segmentation tasks and imaging modalities.
\section{Related Work}

\paragraph{Monte Carlo Dropout} is a technique used by early approaches for uncertainty estimation in image segmentation, which use dropout \cite{srivastava2014dropout} over spatial features to induce probability distributions over the models' outputs \cite{kendall2015bayesian,kendall2017uncertainties}, allowing the drawing of multiple samples at test-time. However, these methods quantify uncertainty pixel-wise, leading them to produce spatially inconsistent segmentation hypotheses. 

\paragraph{Ensembles} \cite{lakshminarayanan2016simple,lee2016stochastic} and \textit{Multi-Head Neural Networks} \cite{ilg2018uncertainty,lee2015m,rupprecht2017learning} are simple methods to produce plausible and consistent output hypotheses. While they may not be able to capture diversity and learn rare variants when ensemble members and network heads are trained independently, that can be circumvented by joint training on an oracle loss \cite{guzman2012multiple}, which only accounts for the lowest-error prediction. The main disadvantages of these approaches are their poor scaling with the number of hypotheses and the latter's requirement to be set at training time.

\paragraph{Variational Bayesian Inference} methods, like the sPU-Net \cite{kohl2018probabilistic}, HPU-Net \cite{kohl2019hierarchical} and PHiSeg \cite{baumgartner2019phiseg} combine conditional variational autoencoders \cite{kingma2014semi,kingma2013auto,rezende2014stochastic,sohn2015learning} with U-Net-based \cite{ronneberger2015u} networks to model the distribution of segmentations given an input image. Input images are encoded into multivariate normal latent spaces that the decoder samples at test-time to produce arbitrarily complex and diverse segmentation hypotheses. However, this approach requires a training-only posterior network, and the placement of the latent variables within the model entails a partial forward pass for each output hypothesis. Recent work addresses these issues with a more constrained low-rank multivariate normal distribution over the logit space, which avoids the use of a posterior network and allows efficient sampling without compromising performance \cite{monteiro2020stochastic}. Other work \cite{hu2019supervised} extends the sPU-Net using variational dropout \cite{kingma2015variational} to predict epistemic uncertainty, and intergrader variability as a target for supervised aleatoric uncertainty estimation. 
\section{Method}

\subsection{Motivation}\label{sec:Motivation}

For many years, due to optimization difficulties and lack of computing power, it was very difficult to train deep neural networks. However, in the last few years, better hardware and new architectural components, such as batch normalization \cite{ioffe2015batch} and residual connections \cite{he2016deep}, have enabled training increasingly deeper and wider networks \cite{he2016deep,huang2017densely,krizhevsky2012imagenet,simonyan2014very,szegedy2015going,tan2019efficientnet}, which achieve high performance in a wide range of tasks. However, unlike their shallower and less accurate counterparts from the past, like the LeNet \cite{lecun1998gradient}, modern neural networks are poorly calibrated, leading to a situation where the probabilities they assign to classes do not reflect their real likelihoods. Though a set of factors such as model capacity, batch normalization and lack of regularization have been put forward as possible causes for miscalibration, the use of hard labels is probably one of the causes at the heart of the problem. When training neural networks to make their predictions match a set of hard labels, which are often the only available ones, it is unreasonable to interpret them as probabilistic models and expect them to output well-calibrated confidence values.

A simple approach to address this issue would be to use soft labels conveying information about real class likelihood. These would not only allow modeling the uncertainty inherent to the data, but would also be likely to enable faster and more data-efficient training. As noted in seminal work on knowledge distillation by Hinton et al. \cite{hinton2015distilling}, compared to hard targets, high entropy soft targets provide much more information per training case and much less variance in the gradient across samples, allowing models to be trained on much fewer data and with significantly higher learning rates. In fact, even noisy soft labels can be of great value, as showcased by recent research on semi-supervised learning \cite{pham2021meta,xie2020selftraining}. 

\subsection{Proposed Method}\label{sec:ProposedMethod}

We propose to use soft labels built from multiple annotations to model uncertainty and address network calibration. Given a set of labels for an image, we average them to produce probabilistic ground-truth masks. Beyond expressing real physicians' uncertainty about zones of interest and their contours, these high entropy soft labels enable our models to enjoy the advantages pointed out by Hinton et al. \cite{hinton2015distilling}. Additionally, note that, for binary variables, the variance can be easily obtained from the mean\footnote{$X = X^2 \implies \mathbf{V}(X) = \mathbf{E}(X^2) - \mathbf{E}(X)^2 = \mathbf{E}(X) - \mathbf{E}(X)^2$.}. Thuerefore, although it can be used as an auxiliary supervision signal, it does not need to be predicted directly by the models.

Depending on the number of annotations per image, the set $S$ of possible ground-truth probabilities will be more or less granular. Formally, let $N$ be the number of annotations, then $S=\{\frac{i}{N}: i \in \mathbb{N} \wedge i \leq N\}$. More annotations per image result in smoother and less noisy ground-truth, which, intuitively, should allow better segmentation performance and uncertainty modeling.

Following Hinton et al. \cite{hinton2015distilling}, we minimize the cross-entropy between the probabilities predicted by the model, $p$, and the ground-truth soft targets, $g$. Note that the dice loss (DL) function, commonly used in segmentation tasks, is not suitable to be used with soft targets. Let $C$ be the number of classes and $N$ the total number of pixels. DL can be defined as
\begin{equation}\label{eq:dice_loss}
    \text{DL}(p, g) = 1 - 2 \frac{\sum_{c=1}^C\sum_{i=1}^N [g_{ci}p_{ci}]}{\sum_{c=1}^C\sum_{i=1}^N [g_{ci} + p_{ci}]}.
\end{equation}
Without loss of generalization, consider the binary classification of a single-pixel image. For $g > 0$, DL is a monotonically decreasing function of $p$. Hence, for $p \in [0, 1]$, the minimum DL will be obtained for $p = 1$.

Consequently, the model is encouraged to binarize its outputs and does not learn to predict uncertainty. This problem could be mitigated by measuring DL at multiple confidence thresholds. However, it would require defining the optimal number of thresholds and their values. Thus, we opt for the more principled cross-entropy loss. Researching overlap-based loss functions that, unlike DL, can be used to match soft targets is a possible direction for future work.

\subsection{Model Architecture}\label{sec:ModelArchitecture}

\begin{figure}[ht]
    \centering
    \includegraphics[width=0.54\linewidth]{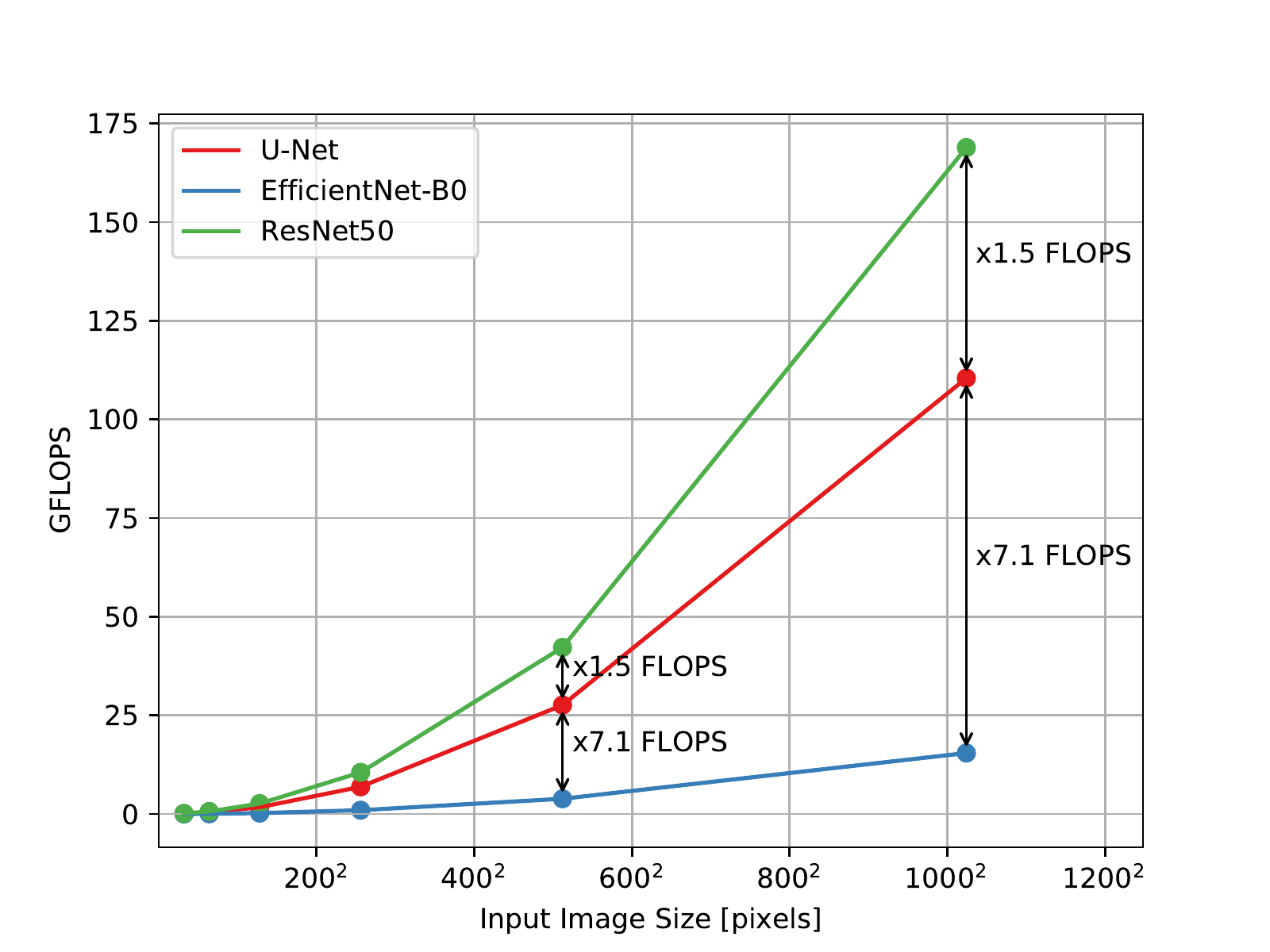}
    \caption{Compute as a function of input image size in pixels, for a U-Net with 16 channel feature maps at the highest resolution using its default encoder (in red), an EfficientNet-B0 (in blue) and a ResNet-50 (in green). An EfficientNet-B0 encoder requires about 7.1 times less FLOPS than the U-Net's default encoder, which in turn requires about 1.5 times less FLOPS than a ResNet-50.}
    \label{fig:encoder_comp}
\end{figure}

We conduct our experiments using a U-Net decoder \cite{ronneberger2015u} with 16-channel feature maps at the highest resolution level. As encoder, we use an EfficientNet-B0 \cite{tan2019efficientnet}. Besides being a better feature extractor than the default U-Net encoder, the EfficientNet-B0 makes for segmentation models whose compute scales better with input image size (see Figure \ref{fig:encoder_comp}), even when compared to other popular encoders \cite{silva2021encoder}, which can be of great value when segmenting high pixel-count medical images. In informal experiments we observed that increasing model capacity within the EfficientNet family did not enhance performance significantly.

\section{Experimental Setup}

\subsection{Datasets and Data Augmentation}

We evaluated our method on datasets from the MICCAI 2021 QUBIQ challenge\footnote{Challenge information and datasets available \href{https://qubiq21.grand-challenge.org/}{at this https url}.}, composed of CTs and MRIs with multiple annotations per case. Except for the four-channel brain tumor MRIs, all images are single-channel. Due to architectural constraints, images are cropped during training to ensure their dimensions are multiples of 32. Crop sizes were set empirically to balance segmentation performance and training time. Table \ref{tab:datasets} summarizes the details of each dataset.

\setlength{\tabcolsep}{6pt}
\begin{table}[ht]
\centering
\caption{Summary of the MICCAI 2021 QUBIQ challenge datasets used.}
\label{tab:datasets}
\resizebox{\textwidth}{!}{%
\begin{tabular}{@{}lccccccc@{}}
\toprule
\multirow{2}{*}{Dataset} &
  \multirow{2}{*}{Modality} &
  \multirow{2}{*}{Tasks} &
  \multirow{2}{*}{Annotators} &
  \multicolumn{2}{c}{Size} &
  \multicolumn{2}{c}{Cases} \\ \cmidrule(l){5-8} 
                  &        &   &   & Slice                                & Crop                     & Train & Validation \\ \midrule
Brain Growth      & 2D MRI & 1 & 7 & $256^2$                     & $256^2$         & 34    & 5          \\
Brain Tumor       & 2D MRI & 3 & 3 & $240^2$                     & $224^2$          & 28    & 4          \\
Kidney            & 2D CT  & 1 & 3 & $497^2$                     & $320^2$         & 20    & 4          \\
\multirow{2}{*}{Prostate} &
  \multirow{2}{*}{2D MRI} &
  \multirow{2}{*}{2} &
  \multirow{2}{*}{6\textsuperscript{$\dagger$}} &
  $640^2$ &
  \multirow{2}{*}{$480^2$} &
  \multirow{2}{*}{48} &
  \multirow{2}{*}{7} \\
                  &        &   &   & \multicolumn{1}{l}{$640 \times 960$} &                          &       &            \\ \bottomrule
\end{tabular}
}
\end{table}

Data augmentation is performed online and consists of the following sequentially applied random transformations: 1) -10\% to 10\% horizontal and vertical translation; 2) $-15\degree$ to $15\degree$ rotation; 3) -10\% to 10\% zoom; 3) horizontal flip with 50\% probability; 4) vertical flip with 0\% probability for the kidney dataset and 50\% for the remaining. Following Fort et al. \cite{fort2021drawing}, we draw multiple augmentation samples per image in a growing batch regime. However, unlike them, we keep the original images in the batch, since we observed this improves performance slightly. Specifically, we augment each batch with three transformations of each image.
\customfootnotetext{$\dagger$}{Except for three cases, which only have 5 annotations.}
\subsection{Evaluation Metrics}

Segmentation is a spatially structured prediction task. Therefore, segmentation models' calibration must be assessed using metrics that take spatial structure into account. Hence, instead of common pixel-wise calibration metrics \cite{brier1950verification,friedman2001elements,naeini2015obtaining}, we measure overlap and surface distance at multiple confidence thresholds. Additionally, following previous work \cite{kohl2019hierarchical,kohl2018probabilistic,monteiro2020stochastic}, we use the generalized energy distance to measure the statistical distance between the ground-truth masks and our models' predictions at the 50\% confidence threshold. Below, we briefly describe the more domain-specific metrics, which may be unknown to some readers. In addition to those, we also measure precision and recall.

\paragraph{Dice Similarity Coefficient (DSC) and Intersection over Union (IoU)} are measures of the overlap between two segmentations. DSC is the ratio between the area of overlap and the sum of the two areas, and is equal to $1 - \text{DL}$ (see Eq. \ref{eq:dice_loss}). IoU is the ratio between the overlap and union areas. When both segmentation masks are empty, we set DSC and IoU to 1.

\paragraph{95\% Hausdorff Distance (95\% HD).} Given two sets of points - two segmentation masks, in this context -, the Hausdorff distance is the maximum distance from a point in one set to the closest point in the other set. The 95\% HD disregards the 5\% most distant pairs of points, ignoring outliers, but still providing a measure of the longest distance between the two sets of points.

\paragraph{Generalized Energy Distance ($D_{\text{GED}}$)} measures the statistical distance between probability distributions. As long as the function $d(\cdot,\cdot)$ is a metric, so is $D_{\text{GED}}$. Following previous work \cite{kohl2019hierarchical,kohl2018probabilistic,monteiro2020stochastic} we define $d(x,y) = 1 - \text{IoU}(x,y)$, which has been proven to be a metric \cite{kosub2019note,lipkus1999proof}. Given the distributions of ground-truth segmentations, $p$, and predicted segmentations, $\hat{p}$, $D^2_{\text{GED}}$ is defined by
\begin{equation}
    D_{\text{GED}}^2(p, \hat{p}) = 2\mathbb{E}_{y\sim p,\hat{y}\sim\hat{p}}[d(y,\hat{y})] - \mathbb{E}_{y,y\sp{\prime}\sim p}[d(y,y\sp{\prime})] - \mathbb{E}_{\hat{y},\hat{y}\sp{\prime}\sim\hat{p}}[d(\hat{y},\hat{y}\sp{\prime})].
\end{equation}
As our models are deterministic, the formula above can be simplified to 
\begin{equation}\label{eq:simple_ged}
    D_{\text{GED}}^2(p, \hat{p}) = 2\mathbb{E}_{y\sim p}[d(y,\hat{y})] - \mathbb{E}_{y,y\sp{\prime}\sim p}[d(y,y\sp{\prime})],
\end{equation}
where $\hat{y}$ is the predicted segmentation mask. The first term of Eq. \ref{eq:simple_ged} is the average distance between predicted and ground-truth annotations, and the second can be interpreted as a measure of ground-truth segmentation diversity.

\subsection{Implementation and Training Details}

We use encoders pre-trained on ImageNet \cite{deng2009imagenet}. Decoder hidden and output layers are initialized using Kaiming \cite{he2015delving} and Xavier initialization \cite{glorot2010understanding}, respectively. Models are trained for 150 epochs - 180 for the $3^{rd}$ brain tumor task -, using batches of 8 images. As optimizer, we use Adam \cite{kingma2014adam}, with $\beta_1=0.9$, $\beta_2=0.999$ and no weight decay. Learning rates are initialized at $10^{-2}$ and decreased to $10^{-4}$ using a cosine annealing schedule \cite{loshchilov2016sgdr}. To ensure reproducibility, we trained and tested each model three times, obtaining similar results across all runs. All experiments are conducted using public PyTorch implementations \cite{Yakubovskiy:2019} under an MIT license. 
\section{Results}

To assess calibration, we start by measuring DSC, precision, recall and 95\% HD between model predictions and averaged ground-truth masks at multiple confidence thresholds. Specifically, we use thresholds ranging from 10\% to 90\% confidence, with a step size of 10\%. 

The averages of these metrics across thresholds are reported in Table \ref{tab:global_results}. Note that the results should be interpreted taking into account image sizes, reported in Table \ref{tab:datasets}, and ground-truth area to image size ratios. For example, the prostate tasks' regions of interest are relatively large, making them easy to overlap and leading to a high DSC. However, the large size of the images - $640 \times 640$ and $640 \times 960$ - and structures leads to apparently high 95\% HDs, compared to those of other tasks. On the other hand, the tiny structures in the third brain tumor segmentation task are challenging to detect and segment, hence the relatively low DSC and recall. Nevertheless, the low image size and object make low 95\% HDs relatively easy to achieve. 

Overall, apart from the second brain tumor segmentation task, which we discuss in more detail below, our method achieves high segmentation performance in all the remaining tasks. Furthermore, the low standard deviations indicate that performance is consistent across multiple confidence thresholds and, therefore, that models are well-calibrated.

\setlength{\tabcolsep}{6pt}
\begin{table}[]
\centering
\caption{Dice score, precision, recall, 95\% Hausdorff distance and ground-truth area to image size ratio, averaged across confidence thresholds ranging from 10\% to 90\%, with a step size of 10\%. Results presented as mean $\pm$ standard deviation.}
\label{tab:global_results}
\resizebox{\textwidth}{!}{%
\begin{tabular}{@{}lccccc@{}}
Task          & Dice Score [\%]   & Precision [\%]    & Recall [\%]       & 95\% HD [pixels] & $\frac{\text{Ground-Truth Area}}{\text{Image Size}}$ [\%]\\ \midrule
Brain Growth  & $93.19 \pm 1.83$  & $93.33 \pm 2.55$  & $93.16 \pm 1.84$  & $3.53 \pm 0.73$ & $8.33 \pm 1.57$                            \\
Brain Tumor 1 & $92.90 \pm 1.09$  & $93.79 \pm 1.01$  & $89.04 \pm 2.30$  &  $6.84 \pm 2.37$ & $3.50 \pm 0.31$                               \\
Brain Tumor 2 & $65.06 \pm 20.27$ & $67.18 \pm 21.75$ & $63.74 \pm 19.57$ & $15.15 \pm 13.18$ & $1.33 \pm 1.47$                               \\
Brain Tumor 3 & $85.73 \pm 4.88$  & $97.67 \pm 1.78$  & $78.34 \pm 6.67$  & $2.27 \pm 2.15$ & $0.29 \pm 0.04$                              \\
Kidney        & $96.04 \pm 1.43$  & $96.57 \pm 1.64$  & $95.65 \pm 2.53$  & $7.67 \pm 3.65$ & $1.90 \pm 0.12$                               \\
Prostate 1    & $95.64 \pm 0.04$  & $94.13 \pm 1.05$  & $97.43 \pm 0.78$  & $16.18 \pm 7.58$ & $8.87 \pm 0.68$                               \\
Prostate 2    & $93.56 \pm 4.58$  & $91.78 \pm 4.09$  & $95.70 \pm 5.32$  & $12.48 \pm 5.46$ & $5.49 \pm 0.58$                               \\ \bottomrule
\end{tabular}%
}
\end{table}

To further assess calibration and uncertainty modeling, we follow previous work \cite{kohl2019hierarchical,kohl2018probabilistic,monteiro2020stochastic} and measure the generalized energy distance ($D_{\text{GED}}$) between the multiple ground-truth annotations and the models' predictions at the 50\% confidence threshold. Additionally, we measure the expected value of the DSC between predictions at 50\% confidence and each physician's annotations - which is not equivalent to measuring the DSC between model predictions and average ground-truth masks at the same threshold. Results are reported in Table \ref{tab:ged}.

Except for the second brain tumor segmentation task, the remaining tasks' $D^2_{\text{GED}}$ is very low, meaning the models' predictions closely match the distributions of ground-truth annotations. In fact, in most cases, the expected value of the IoU distance between model predictions and ground-truth annotations is lower than that of the IoU distance between annotations by different physicians, indicating that, on average, our models do a better task at matching a physician's annotations than other physicians do, which is remarkable, especially considering the small dimension of the datasets, composed of 20 to 48 samples.

The overall worse performances are registered for the second and third brain tumor segmentation tasks. For the latter, the lower performance is largely justified by the difficulty of segmenting its tiny structures. However, in the former case, the difficulty lies in the high variability between ground-truth masks. Even though three annotators may not be enough to represent all the segmentation hypotheses in this task, we suspect the annotations from one of the physicians to be incorrect, as their average IoU distance to the others is 94.15\%, and the distance between the other physicians' annotations is only 19.82\%.  

Finally, note that the expected value of the DSC between predictions at 50\% confidence and each physician's annotations is generally high, meaning that beyond matching the averaged predictions of multiple physicians, the masks produced by our models also match individual physicians' annotations well.

\setlength{\tabcolsep}{6pt}
\begin{table}[ht]
\centering
\caption{From the $2^{nd}$ to the $5^{th}$ column: squared generalized energy distance; expected IoU distance between predictions and ground-truth masks; ground-truth diversity; expected DSC between predictions at 50\% confidence and each physician's annotations. Results presented as mean $\pm$ standard deviation.}
\label{tab:ged}
\resizebox{\textwidth}{!}{%
\begin{tabular}{@{}lcccc@{}}
\toprule
Task & $D^2_{\text{GED}}$ & $\mathbb{E}_{y\sim p}[1-\text{IoU}(y,\hat{y})]$ & $\mathbb{E}_{y,y\sp{\prime}\sim p}[1-\text{IoU}(y,y\sp{\prime})]$ & $\mathbb{E}_{y\sim p}[\text{DSC}(y,\hat{y})]$ \\ \midrule
Brain Growth  & $0.1323 \pm 0.0077$ & $0.1876 \pm 0.0087$ & $0.2429 \pm 0.0124$ & $89.63 \pm 01.60$ \\
Brain Tumor 1 & $0.1455 \pm 0.0622$ & $0.1393 \pm 0.0492$ & $0.1330 \pm 0.0497$ & $92.39 \pm 03.78$ \\
Brain Tumor 2 & $0.6731 \pm 0.5631$ & $0.6843 \pm 0.3167$ & $0.6955 \pm 0.0835$ & $34.27 \pm 43.45$ \\
Brain Tumor 3 & $0.2515 \pm 0.1928$ & $0.2272 \pm 0.1523$ & $0.2030 \pm 0.1306$ & $86.25 \pm 10.22$ \\
Kidney        & $0.0613 \pm 0.0077$ & $0.0814 \pm 0.0105$ & $0.1015 \pm 0.0150$ & $95.73 \pm 01.59$ \\
Prostate 1    & $0.0950 \pm 0.0692$ & $0.1096 \pm 0.0569$ & $0.1242 \pm 0.0478$ & $94.07 \pm 03.80$ \\
Prostate 2    & $0.0988 \pm 0.0907$ & $0.1431 \pm 0.0679$ & $0.1874 \pm 0.0828$ & $90.99 \pm 15.25$ \\ \bottomrule
\end{tabular}%
}
\end{table}
\section{Discussion}

We proposed a new way of approaching uncertainty modeling in image segmentation. Instead of building models that learn independently from the annotations of multiple physicians and can produce multiple segmentation hypotheses for a given image, we train deliberately deterministic models on the joint predictions of physician ensembles, using the averages of their predictions as soft labels. 

We evaluated our method on datasets from the MICCAI 2021 QUBIQ challenge, showing that it results in well-calibrated models that, on average, match physicians' predictions better than other physicians. The results show that our system exhibits a good performance on this task, competitive with other approaches.

In future work, we plan to test this technique on larger datasets and more challenging tasks with multiple classes, possibly including problems outside the scope of medical image segmentation. Additionally, we intend to investigate if the soft labels used in our method allow the more data-efficient training generic soft labels do \cite{hinton2015distilling}. Finally, given soft labels' role in recent teacher-student semi-supervised learning methods \cite{xie2020selftraining,pham2021meta}, we plan to assess if networks trained on soft labels, like ours, can be better teachers than those trained on hard labels.
\section*{Acknowledgments}

This work was supported by national funds through Fundação para a Ciência e Tecnologia (FCT), under the project with reference UIDB/50021/2020.

\bibliographystyle{splncs04}
\bibliography{references.bib}
\end{document}